\pdfoutput=1

\documentclass[11pt]{article}

\usepackage[]{acl}

\usepackage{times}
\usepackage{latexsym}
\usepackage{soul}

\usepackage[T1]{fontenc}

\usepackage[utf8]{inputenc}

\usepackage{microtype}

\usepackage{inconsolata}

\usepackage{graphicx} 
\usepackage{natbib}  
\usepackage{caption} 
\usepackage{multirow}
\usepackage{algorithm}
\usepackage{colortbl}
\usepackage{algpseudocode}
\usepackage{amsmath}
\usepackage{booktabs}
\usepackage{amsfonts}
\usepackage{blkarray, bigstrut}
\parskip=\medskipamount
\usepackage{amssymb}
\usepackage{pifont}
\newcommand{\cmark}{\ding{51}}%
\newcommand{\xmark}{\ding{55}}%

\newcommand{\xlopsummm}[1]{{\textsc{Xl-OpSumm(#1)}}}

\usepackage[labelfont=bf]{caption}
\usepackage{multirow}
\newcommand*\rot{\rotatebox{90}}
\usepackage{newfloat}
\usepackage{listings}

\usepackage{xcolor}
\usepackage[capitalise,noabbrev]{cleveref}
\usepackage{hyperref}

\newcommand{\nonumberfootnote}[1]{
    \begingroup
    \let\thefootnote\relax
    \footnotetext{#1}
    \endgroup
}
\newcommand\blfootnote[1]{%
  \begingroup
  \renewcommand\thefootnote{}\footnote{#1}%
  \addtocounter{footnote}{-1}%
  \endgroup
}

\title{Distilling Opinions at Scale: Incremental Opinion Summarization using \textsc{Xl-OpSumm}}

\author{Sri Raghava Muddu$^\dag$$^\clubsuit$,
Rupasai Rangaraju$^\dag$$^\clubsuit$, Tejpalsingh Siledar$^\dag$$^\clubsuit$ , Swaroop Nath$^\dag$$^\clubsuit$ , 
\\ 
\textbf{Pushpak Bhattacharyya$^\clubsuit$, Swaprava Nath$^\clubsuit$, } 
\\
\textbf{Suman Banerjee$^\spadesuit$, Amey Patil$^\spadesuit$, Muthusamy Chelliah$^\spadesuit$,}\\
\textbf{Sudhanshu Shekhar Singh$^\spadesuit$, Nikesh Garera$^\spadesuit$}\\
        $^\clubsuit$Computer Science and Engineering, IIT Bombay, India, \\
        $^\spadesuit$Flipkart, India \\
        \texttt{\{sriraghava,rupasai,tejpalsingh,swaroopnath,pb,swaprava\}@cse.iitb.ac.in}
        \\
        }

\begin{document}
\maketitle
\blfootnote{$^\dag$ Equal Contribution}
\begin{abstract}
Opinion summarization in e-commerce encapsulates the collective views of numerous users about a product based on their reviews. Typically, a product on an e-commerce platform has thousands of reviews, each review comprising around 10-15 words. While Large Language Models (LLMs) have shown proficiency in summarization tasks, they struggle to handle such a large volume of reviews due to context limitations. To address this, we propose a scalable framework called \textbf{\textsc{Xl-OpSumm}} that generates summaries incrementally with the help of \textsc{Aspect Dictionary} (Refer to Section \ref{sec: framework}). However, the existing test set, AMASUM has only 560 reviews per product on average. Due to the lack of a test set with thousands of reviews, we created a new test set called \textbf{\textsc{Xl-Flipkart}} by gathering data from the Flipkart website and generating summaries using GPT-4\footnote{GPT-4:\href{https://openai.com/gpt-4}{\tt openai/gpt-4}}. Through various automatic evaluations and extensive analysis, we evaluated the framework's efficiency on two datasets, AMASUM and \textsc{Xl-Flipkart}. Experimental results show that our framework, \textsc{Xl-OpSumm} powered by \textsc{Llama-3-8B-8k}, achieves an average ROUGE-1 F1 gain of $\mathbf{4.38\%}$ and a ROUGE-L F1 gain of $\mathbf{3.70\%}$ over the next best-performing model.
\end{abstract}

\section{Introduction}
E-commerce websites are valuable sources of product reviews, aiding users in well-informed purchasing decisions. Yet, sifting through numerous reviews can be daunting and time-consuming. Opinion summarization offers a solution by summarizing the opinions presented in product reviews \citep{hu2006opinion, wang-ling-2016-neural, angelidis-lapata-2018-summarizing, siledar-etal-2023-synthesize}. However, their utility is limited when confronted with the vast number of reviews, typical of e-commerce platforms. Recent advancements in opinion summarization \citep{bhaskar-etal-2023-prompted, hosking-etal-2023-attributable} address this by scaling systems to accommodate a larger number of reviews, yet they still fall short of fully harnessing the vast array of reviews often numbering in the thousands.

Recent studies have demonstrated that Large Language Models (LLMs) can generate effective opinion summaries in zero-shot prompt settings \citep{siledar2024prompt}. However, when dealing with large contexts, LLMs often struggle to retrieve relevant information from the middle of the context \citep{liu2023lost}. Furthermore, despite their ability to process a large number of tokens, LLMs are constrained by context limits and cannot accommodate the entire set of reviews, which typically number in the thousands. 

To address this issue, incremental and hierarchical approaches have been proposed by \citet{chang2023booookscore}. Nonetheless, these methods may not effectively manage conflicting opinions about specific aspects across different chunks of reviews while updating the summary.

The unavailability of any large-scale (ranging in thousands of reviews) test sets hinders progress in this area. To address these issues, we first create \textsc{Xl-Flipkart}, a test set containing $\sim3680$ reviews on average per product for $25$ products from the Flipkart Website\footnote{Flipkart: \href{https://www.flipkart.com}{\tt flipkart.com}}. We employ GPT-$4$ to annotate summaries \citep{Gilardi_2023, Huang_2023, siledar2024product}. Next, we propose using an incremental approach to summarize reviews and generate summaries. This we claim has two benefits: (\textit{a}) in the presence of a fresh set of reviews, after a certain period of time (usually the case in the e-commerce domain), our approach emerges as an efficient way of updating summaries, and (\textit{b}) does not face context-limit issues which is usually the case when handling such large amount of reviews.

\begin{figure*}[htp]
    \centering
    \includegraphics[width=2\columnwidth]{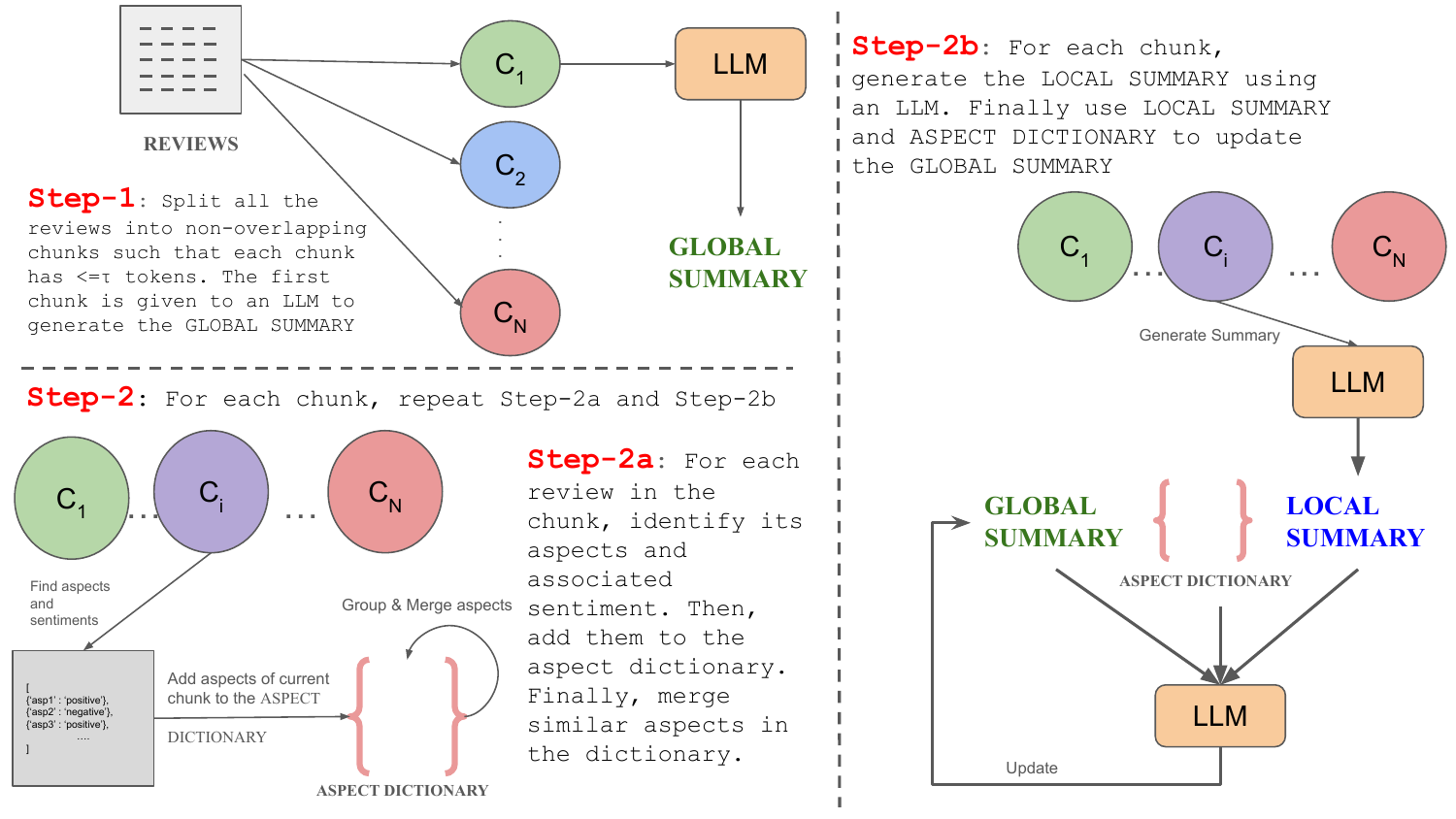}
    \caption{Illustration of our \textbf{\textsc{Xl-OpSumm}} framework. First, reviews are divided into non-overlapping chunks based on threshold. Then for each chunk, the \textsc{Aspect dictionary} is updated, the \textsc{Local Summary} is generated and the \textsc{Global Summary} is updated as shown above. Refer to the section \ref{sec: framework} for more details about this framework }
    \label{fig:overview}
\end{figure*}

\noindent Our contributions are:
\begin{enumerate}
    \item \textbf{\textsc{Xl-Flipkart}}, a large-scale ($\mathbf{\sim3600}$ reviews on average per product) test set of $25$ products gathered from the Flipkart website annotated using GPT-$4$ (Section \ref{sec: testset}). \textit{To the best of our knowledge}, this is the first large-scale opinion summarization test set.
    \item \textbf{\textsc{Xl-OpSumm}}, a large-scale opinion summarization framework that uses an incremental approach capable of generating summaries efficiently using thousands of reviews without any context limitation (Figure \ref{fig:overview}, Section \ref{sec: framework}). Experimental demonstrations indicate that our \textsc{Xl-OpSumm} framework powered by \textsc{Llama-3-8B-8k}, achieves an average ROUGE-1 F1 gain of $\mathbf{4.38\%}$ and a ROUGE-L F1 gain of $\mathbf{3.70\%}$ over the next best-performing model (Table \ref{Table: automatic_eval_results}).
    \item Qualitative and comparative analysis indicating the efficacy of our \textsc{Xl-OpSumm} framework in handling thousands of reviews for generating comprehensive opinion summaries compared to existing approaches (Sections \ref{sec: qualitative} \& \ref{sec: comparative}).
\end{enumerate}

\section{Related Work}
\textbf{Opinion Summarization} employs two main approaches: extractive and abstractive. Extractive methods involve selecting the most pertinent sentences directly from the input text, while abstractive techniques generate a condensed version of the opinions expressed.
\\
A Widely used extractive method is the centroid approach, which ranks sentences by relevance to the input text. Another technique is clustering, where sentences are grouped by themes and representative ones are chosen from each cluster. Centroid-based methods include \citep{Radev2004CentroidbasedSO, rossiello-etal-2017-centroid, gholipour-ghalandari-2017-revisiting}, which prioritize sentence selection based on their centrality to the input, and graph-based methods \citep{Erkan2004LexRankGL, mihalcea-tarau-2004-textrank, zheng-lapata-2019-sentence}, which construct graphical representations of the text and extract sentences located at central nodes.
\\ Abstractive opinion summarization is often performed in a self-supervised manner by treating a single review as a pseudo-summary. Various approaches exist for selecting pseudo-summaries and their corresponding input reviews.  \citet{brazinskas-etal-2020-unsupervised} employed a random selection of N reviews per entity to construct N pseudo-summary, review pairs.  \citet{amplayo-lapata-2020-unsupervised} sampled a review randomly and generated noisy versions of it as input reviews.  \citet{Amplayo2020UnsupervisedOS} used aspect and sentiment distributions to guide pseudo-summary sampling.  \citet{elsahar-etal-2021-self} selected input reviews with high TF-IDF cosine similarity to a randomly sampled pseudo-summary. \citet{wang-wan-2021-transsum} focused on reducing opinion redundancy by learning aspects and sentiment embeddings to generate highly relevant review-pseudo-summary pairs.  \citet{im-etal-2021-self} used a synthetic dataset creation strategy similar to \citet{brazinskas-etal-2020-unsupervised}, extending it to multimodal data.  \citet{Ke2022ConsistSumUO} emphasized consistency of aspects and sentiment between reviews and pseudo-summary by using constrained sampling.  Finally, \citet{siledar-etal-2023-synthesize} leveraged lexical and semantic similarities for creating synthetic datasets and \citet{siledar2024product} uses additional information sources such as product description and question answers of a product to create the synthetic dataset. However, these methods fail to accommodate a substantial volume of review sets as they typically rely on a limited number of input reviews (e.g., 10 reviews) to produce the opinion summary.
\\
\\ \textbf{Large Scale Opinion Summarization} Recent opinion summarization systems such as \citep{bhaskar-etal-2023-prompted, hosking-etal-2023-attributable, jiang-etal-2023-large} include a large number of reviews. \citet{bhaskar-etal-2023-prompted} explores prompting by testing \cite{openai2023} and introduces various pipelines whereas \citet{jiang-etal-2023-large} introduced a review sampling strategy that uses sentiment analysis and two-stage training scheme to generate the opinion summary. \citet{hosking-etal-2023-attributable} encodes the reviews into discrete latent space and then generates the summary by decoding the frequent encodings.
\\
\\ \textbf{Incremental Summarization} \cite{chowdhury2024incremental} proposes CoverSumm an algorithm to perform centroid-based extractive opinion summarization incrementally. \cite{chang2023booookscore} uses incremental and hierarchical approaches to summarise book-length text. We propose \textsc{xl-opsumm} framework for a large-scale opinion summarization system that generates the opinion summary incrementally.

\begin{table*}
    \centering
    \resizebox{2\columnwidth}{!}{%
    \begin{tabular}{p{0.6\columnwidth} | p{0.8\columnwidth} | p{0.6\columnwidth} }
    \toprule
        \small{\textbf{Previous \textsc{Global Summary}}} &
        \small{\textbf{\textsc{Local Summary}}} & 
        \small{\textbf{Updated \textsc{Global Summary}}} \\
    \midrule
    \small{The provided text appears to be a collection of customer reviews for the Realme 8 smartphone. Customers have provided a mix of positive and negative feedback on various aspects such as display, fingerprint sensor, camera quality, battery life, performance, and charging speed. \textcolor{red}{Some users have expressed dissatisfaction with the camera quality and overall performance}, while others praised the phone for its display, fingerprint sensor, battery backup, and value for money. \textcolor{red}{It's evident that while the Realme 8 has received some positive feedback, there are also concerns that potential buyers should consider}.}
    &
    \small{Realme 8 smartphone \textcolor{blue}{offers sturdy and strong build quality}, although its back is prone to fingerprints, necessitating the use of a back case. \textcolor{red}{The rear camera, powered by Sony IMX sensors, delivers excellent results, earning a 5/5 rating. However, the front camera captures only decent pictures}. \textcolor{red}{The phone's performance is commendable, thanks to the MTK G 95 processor, which smoothly handles day-to-day applications and gaming}. It's well-suited for games like \textcolor{blue}{COD, BGMI, and Fortnite}. Additionally, the impressive 5,000 mAh battery can easily last up to a day with normal usage, and it supports 30 Watts fast charging, allowing for a full charge within an hour. Lastly, the \textcolor{blue}{SUPER AMOLED display enhances the viewing experience, surpassing other LCD or IPS LED displays in the same price range}. Overall, the Realme 8 is a reliable choice for those seeking a value-for-money smartphone in this category.} 
    &
    \small{This review highlights the Realme 8 smartphone's \hl{performance}, \hl{camera quality}, battery life, and display. The user \hl{praises the phone's sturdy build}, powerful back camera, seamless performance for gaming, and impressive battery life. They also appreciate the \hl{SUPER AMOLED display, which outshines other IPS LED displays in the same price range}. However, they note that the \hl{front camera's quality is average}, and they experience a {\hl{slight battery degradation after three days}}. Despite these minor drawbacks, the overall sentiment is positive, and the user recommends the phone based on its value for money and great features. {Overall, the Realme 8 receives a rating of 4.5 out of 5 stars}.}
    
    \\
        \bottomrule
    \end{tabular}%
    }
    \caption{\textbf{Qualitative Analysis.} This table demonstrates how our framework updates summaries. The \textsc{Local Summary} represents the summary of the current chunk, while the Previous \textsc{Global Summary} encapsulates the summaries of all previously processed chunks. The Updated \textsc{Global Summary} combines the summaries of all chunks up to and including the current chunk. Conflicting aspect opinions between the Local and Global summaries are shown in red, and new aspects are highlighted in blue. Updated information using the Aspect Dictionary is marked in yellow. For more details, refer to Section \ref{sec: qualitative}.}
    \label{Table: qualitative}

\end{table*}

\section{\textsc{Xl-OpSumm} Framework}
\label{sec: framework}

To summarize reviews of a product, we split them into non-overlapping chunks, each with up to $\tau$ tokens. We then analyze each chunk using three elements: \textbf{\textsc{Local Summary}}, \textbf{\textsc{Global Summary}}, and \textbf{\textsc{Aspect Dictionary}}. The \textsc{Local Summary} is the summary of all reviews in the current chunk, while the \textsc{Global Summary} is the summary of all previous chunks. The \textsc{Aspect Dictionary} contains aspects and their corresponding positive, negative, and neutral sentiment counts expressed by users from previous segments. Here are the steps as shown in figure \ref{fig:overview} to obtain the final summary for the product:

\noindent \textbf{Step-1}: The \textsc{Global Summary} is initialized with a summary generated by an LLM using all reviews from the first chunk.

\noindent \textbf{Step-2}: For each chunk, we repeat the following procedure: 

\noindent \textbf{Step-2a}: For each review in the chunk, we identify its aspects and corresponding sentiments using the Aspect-Based Sentiment Analyser (ABSA) Model. We then update the \textsc{Aspect Dictionary} by adding the sentiments of aspects in the current chunk to the \textsc{Aspect Dictionary}. To avoid redundancy, we merge similar aspects into a single aspect by encoding the aspect names in the dictionary using Sentence Transformer \citep{reimers-2019-sentence-bert} and clustering them using the fast clustering algorithm. The sentiment counts of all aspects in one cluster are added together and finally represented using one aspect name.

\noindent \textbf{Step-2b}: We use an LLM to generate the summary of the current chunk and assign it to the \textbf{\textsc{Local Summary}}. Finally, we use the \textsc{Aspect Dictionary} and the \textsc{Local Summary} to update the \textsc{Global Summary}. The procedure to update the \textsc{Global Summary} is presented in appendix \ref{sec: GSU}.

After processing all the chunks, the summary in the \textsc{Global Summary} element is considered as the final summary for the product.

\section{Dataset Details}
\begin{table}[htp]
    \centering
    \resizebox{\columnwidth}{!}{%
    \begin{tabular}{lcc}
    \toprule
     &  \textbf{AMASUM} & \textbf{\textsc{Xl-Flipkart}}\\
    \midrule
    Average \#reviews per entity  & 560.43 & 3682.88\\
    Average \#sentences per review & 3.64 & 1.63\\
    Average \#words per sentence & 13.72 & 10.23\\
    \bottomrule
    \end{tabular}%
    }
    \caption{Dataset statistics of AMASUM, \textsc{Xl-Flipkart}}
    \label{Table: stats}
\end{table}%

\noindent\textbf{AMASUM} \cite{brazinskas-etal-2021-learning} involves the summarization of reviews of various products from Amazon website\footnote{Amazon: \href{https://www.amazon.in/}{\tt amazon.in}}, averaging over 560 reviews per product. In the original dataset, references are categorized into 'verdict', 'pros', and 'cons'. Following \citet{hosking-etal-2023-attributable}, we merge them to form unified summaries. We then narrowed down the original dataset to four prevalent categories (Electronics, Shoes, Sports \& Outdoors, Home \& Kitchen) and sampled a subset of 50 entities, resulting in a total of 200 products. Various statistics of the test set are recorded in Table \ref{Table: stats}.


\section{Testset Creation: \textsc{Xl-Flipkart}}
\label{sec: testset}
The existing AMASUM test set contains approximately $560$ reviews per product. However, in a real e-commerce environment, the number of reviews per product typically reaches into the thousands, which is not represented by the AMASUM dataset. To evaluate our \textsc{Xl-OpSumm} framework in a context closer to real-world scenarios, we collected reviews of $25$ mobile products from the Flipkart website. As shown in Table \ref{Table: stats}, each product in this dataset has around $3,680$ reviews on average. This number is nearly $6.5$ times greater than the average number of reviews per product in the AMASUM dataset. 

Generating summaries for such a large volume of reviews is not only time-consuming but also very challenging for humans. Based on studies by \citet{siledar2024product} which indicate that humans prefer GPT-generated summaries over those written by humans, we utilized GPT-4-turbo to generate the summaries for the products we collected. The prompt used for generating these summaries with GPT-4-turbo is provided below.

\hangindent=2em
\hangafter=1
\noindent\textbf{Prompt}: \texttt{Following are the reviews for a
product. Generate a summary of the opinions
as a review itself with a word limit of under
100 words. Use information from the given
reviews only to generate the summary. \\
\textbf{reviews: [r1,...,rk]}}

\section{Experiments}
\subsection{Baseline Models}
\label{sec: baselines}
We evaluate our framework against various baselines, including both abstractive and extractive systems. Important recent state-of-the-art work is mentioned in this section. Refer to the Appendix \ref{sec: other_baselines} for all the other baselines we considered for this work.
\subsubsection{Non-LLM Baselines}
We evaluated our framework against the following Non-LLM Models 
\par \textbf{HERCULES\textsubscript{EXT}} \cite{hosking-etal-2023-attributable} computes extractive summaries by calculating te centroid from each evidence set generated by using HERCULES based on ROUGE-2 F1 score.\\
\par \textbf{BiMeanVAE and COOP} \cite{iso-etal-2022-comparative} work by encoding entire reviews into continuous latent vectors. BiMeanVAE takes the average of these encodings while COOP calculates the optimized combination of review encodings.
\par \textbf{\textsc{HERCULES}\textsubscript{ABS}} \cite{hosking-etal-2023-attributable} represents a method that aggregates reviews into summaries by identifying frequent opinions in discrete latent space.
\subsubsection{LLM Baselines} 
We evaluated our framework against the following LLM Models 
\par \textbf{\textsc{Llama-3-8B-8k}}\footnote{Llama-3:\href{https://llama.meta.com/llama3/}{\tt meta/llama-3}} is an open source large language model with 8B parameters and 8k context limit. 
\par \textbf{\textsc{Phi-3-Mini-3.8B-4k}} \cite{abdin2024phi} is an open source 3.8B parameter model with 4k context limit 
\par \textbf{\textsc{Phi-3-Mini-3.8B-128k}} \cite{abdin2024phi} is an open source 3.8B parameter model with 128k context limit
\par \textbf{\textsc{Llama-3-8B-8k-Incremental}} is a method to update the existing summary incrementally using a chunk of reviews \cite{chang2023booookscore} with the help of \textsc{Llama-3-8B-8k} model.
\par \textbf{\textsc{Llama-3-8B-8k-Hierarchical}} is a method of summarizing chunks of reviews and then hierarchically merging the summaries until one summary \cite{chang2023booookscore} using the \textsc{Llama-3-8B-8k} model.
\par \textbf{\textsc{Phi-3-Mini-3.8B-128k-Incremental}} is a method to update the existing summary incrementally using a chunk of reviews \cite{chang2023booookscore} with the help of \textsc{Phi-3-Mini-3.8B-128k} model.
\par \textbf{\textsc{Phi-3-Mini-3.8B-128k-Hierarchical}} is a method of summarizing chunks of reviews and then hierarchically merging the summaries until one summary \cite{chang2023booookscore} using the \textsc{Phi-3-Mini-3.8B-128k} model.

\subsection{Implementation Details}
We conducted all experiments using Nvidia DGX A100 GPUs with 80GB of memory. For the large language models (LLMs) used in our experiments, we set the temperature to 0.8. To populate the aspect dictionary, we employed the Instruct ABSA model \cite{varia-etal-2023-instruction} as our aspect-based sentiment analyzer. Within our framework, we experimented with two LLM options: \textsc{Llama-3-8B-8k} and \textsc{Phi-3-Mini-3.8B-4k} \cite{abdin2024phi}. When using \textsc{Llama-3-8B-8k}, we set the $\tau$ value to 4000, whereas for \textsc{Phi-3-Mini-3.8B-4k}, the $\tau$ value was scaled down to 2700 due to its context limitation.
\section{Results and Analysis}
In this section, we show results on various automatic reference-based metrics and reference-free metrics as well. We also analyze our model's performance qualitatively and comparatively with other models' summaries. 
\begin{table*}[h]
    \centering
    \small
    \resizebox{2\columnwidth}{!}{%
    \begin{tabular}{ cclcccccccc}
    \toprule
    &&& \multicolumn{4}{c}{\textbf{AmaSum}} & \multicolumn{4}{c}{\textbf{\textsc{Xl-Flipkart}}} \\
    \cmidrule(lr){4-7}\cmidrule(lr){8-11}
    &\textit{abs?} &\textbf{Model}  & R1 $\uparrow$ & R2 $\uparrow$ & RL $\uparrow$ & BERT-F1 $\uparrow$ & R1 $\uparrow$ & R2 $\uparrow$ & RL $\uparrow$ & BERT-F1 $\uparrow$  \\
    \midrule
    \multirow{5}{*}{\textit{\rot{Extractive}}}&\xmark  & Clustroid    & $17.92$ & $2.13$ & $10.74$ & $84.27$ &0.60&0.1&0.60 & $79.21$ \\
    &\xmark &LexRank   & $22.70$ & $3.10$ & $12.93$ & $83.89$&$9.66$&$0.59$&$6.23$ & $82.62$ \\
    &\xmark &QT    & $21.97$ & $1.66$ & $11.52$  &$83.35$&$18.83$&$1.47$&$10.30$&$81.65$  \\
    &\xmark &SemAE   & $21.31$ & $1.75$ & $11.30$  &$83.32$ &-&-&-&- \\
    &\xmark & HERCULES\textsubscript{EXT}   & $25.49$ & $3.47$ & $12.91$ &$84.01$ &$21.99$&$1.01$&$10.16$ &$82.94$ \\
    \midrule
    \multirow{5}{*}{\textit{\rot{Abstractive}}}&\cmark &CopyCat  &  $16.77$ & $1.57$ & $10.40$ &$83.96$ &-&-&-&-   \\
    &\cmark &BiMeanVAE  &  $22.12$ & $2.23$ & $12.41$  &$83.85$&$8.86^\dagger$&$0.70^\dagger$&$6.20^\dagger$&$82.67^\dagger$  \\
    &\cmark &COOP & $24.63$ & $3.04$ & $\mathbf{14.04}$ &$84.38$ &$9.76^\dagger$&$1.10^\dagger$&$6.71^\dagger$&$82.32^\dagger$\\
    &\cmark &HERCULES\textsubscript{ABS}   & $20.21$ & $2.24$ & $11.72$  &$84.37$ &$17.21$&$0.82$&$9.76$ &$82.88$\\   
    \midrule
    \multirow{5}{*}{\textit{\rot{INC / HIE}}}&\cmark & \textsc{Llama-3-8B-8k-Incremental} &$25.19$ & $\mathbf{3.95}$& $13.35$& $84.30$&$\underline{38.98}$ &$\underline{8.56}$ &$\underline{20.56}$ & $\underline{86.88}$\\
    &\cmark & \textsc{Phi-3-Mini-3.8B-128k-Incremental} &$20.93$&$2.12$&$11.18$&$83.04$&$35.87$&$6.58$&$17.96$&$85.96$ \\
    &\cmark & \textsc{Llama-3-8B-8k-Hierarchical} &$25.07$ & $\underline{3.88}$& $12.73$& $84.08$&$33.16$ &$8.30$ &$17.41$ & $86.70$\\
    &\cmark & \textsc{Phi-3-Mini-3.8B-128k-Hierarchical} &$24.27$&$2.81$&$12.19$&$84.02$&$31.09$&$7.14$&$14.22$&$85.56$ \\
    \midrule
    \multirow{3}{*}{\textit{\rot{LLMs}}}&\cmark & \textsc{Phi-3-Mini-3.8B-4k} & $25.34$& $2.60$& $12.66$& $84.16$&$31.39$ & $5.90$&$14.62$ & $80.99$\\
    &\cmark & \textsc{Phi-3-Mini-3.8B-128k} & $24.14$& $2.56$& $12.63$& $84.36$&$33.82$& $7.77$&$15.62$ & $85.76$ \\
    &\cmark & \textsc{Llama-3-8B-8k} &$\underline{26.13}$ &$3.12$ & $13.51$& $\underline{84.68}$&$35.35$ &$7.56$ & $17.42$&$83.77$ \\
    \midrule
    \multirow{2}{*}{\textit{\rot{Ours}}} &\cmark & \textsc{Xl-OpSumm(Phi-3-Mini-3.8B-4k)} & $24.78$& $2.55$&$12.72$ & $84.59^\star$& ${37.71}^\star$& $6.76$&${17.83}^\star$ &$86.45^\star$ \\
     &\cmark & \textsc{Xl-OpSumm(Llama-3-8B-8k)} & $\mathbf{26.88}^\star$& ${3.52}^\star$& $\underline{13.85}^\star$&$\mathbf{85.11}^\star$ &$\mathbf{39.78}^\star$ &$\mathbf{8.86}$ &$\mathbf{21.31}^\star$ &$\mathbf{87.38}^\star$ \\
    \bottomrule
    \end{tabular}%
    }
    \caption{\textbf{Results on AmaSum, and \textsc{Xl-Flipkart} datasets} . \textit{INC/HIE} indicates that the model uses either an Incremental or a Hierarchical approach. Bold and underlined indicate the best and
second-best scores. $\star$ indicates p-value < 0.05 on \textbf{Wilcoxon Signed-Rank Test} of \textsc{Xl-OpSumm} framework models against their corresponding base LLMs (e.g. \textsc{Xl-OpSumm(Llama-3-8B-8k)} vs \textsc{Llama-3-8B-8k}). $\dagger$ indicated scores obtained by sampling 8 reviews randomly from test set.}
    \label{Table: automatic_eval_results}
\end{table*}%
\subsection{Automatic Evaluation}
\begin{table*}[h]
    \centering
    \small
    \resizebox{1.5\columnwidth}{!}{%
    \begin{tabular}{cllcccc}
    \toprule
    &\multicolumn{1}{c}{\textbf{Model}}&\multicolumn{2}{c}{\textbf{GPT-3.5}} & \multicolumn{2}{c}{\textbf{\textsc{Mistral-7B}}}&  \\
    \cmidrule(lr){3-4}\cmidrule(lr){5-6}
    && FL$\uparrow$ & CO$\uparrow$ & FL$\uparrow$ & CO$\uparrow$ & BooookScore$\uparrow$ \\
    \midrule
    
    \multirow{1}{*}{\textit{\rot{ABS}}}&HERCULES\textsubscript{ABS} &$3.76$&$1.84$&$4.4$&$2.36$& $59.46$\\ \\
    
    \midrule
    \multirow{4}{*}{\textit{\rot{INC/HIE}}}&\textsc{Llama-3-8B-8k-Incremental} &$\underline{4.72}$&$\mathbf{3.72}$&$4.56$&$\underline{4.16}$&$70.19$ \\
    &\textsc{Phi-3-Mini-3.8B-128k-Incremental} &$3.60$&$2.84$&$4.16$&$3.36$& $57.86$  \\
    &\textsc{Llama-3-8B-8k-Hierarchical} &$\mathbf{4.80}$&$3.60$&$\mathbf{4.92}$&$\mathbf{4.44}$&$\underline{71.58}$ \\
    &\textsc{Phi-3-Mini-3.8B-128k-Hierarchical} &$4.36$&$3.48$&$4.64$&$3.92$&$63.12$ \\
    \midrule
    \multirow{3}{*}{\textit{\rot{LLMs}}}&\textsc{Llama-3-8B-8k} &$4.64$&$3.44$&$\underline{4.68}$&$4.08$& $65.06$ \\
    &\textsc{Phi-3-Mini-3.8B-4k} &$4.12$&$3.40$&$4.48$&$3.76$& $58.97$\\ 
    &\textsc{Phi-3-Mini-3.8B-128k} &$4.60$&$3.56$&$4.48$&$4.04$& $61.86$  \\
    \midrule
    \multirow{2}{*}{\textit{\rot{Ours}}}&\textsc{Xl-OpSumm(Phi-3-Mini-3.8B-4k)} &$4.60$&$3.52$&$4.44$&$\mathbf{4.44}$& $61.41$ \\ 
    &\textsc{Xl-OpSumm(Llama-3-8B-8k)} &$4.68$&$\underline{3.64}$&$4.56$&$\mathbf{4.44}$& $\mathbf{85.60}$\\
    \bottomrule
    \end{tabular}%
    }
    \caption{Reference free evaluation on AMASUM Dataset. \textit{INC/HIE} indicates that the model uses either an Incremental or a Hierarchical approach. FL represents the average fluency score across all summaries generated by the model, while CO denotes the average coherency score. Refer to Appendix \ref{sec: metrics} for more description of these metrics.}
    \label{Table: reference_free_amasum}
\end{table*}%

\begin{table*}[h]
    \centering
    \small
    \resizebox{1.5\columnwidth}{!}{%
    \begin{tabular}{cllcccc}
    \toprule
    &\multicolumn{1}{c}{\textbf{Model}}&\multicolumn{2}{c}{\textbf{GPT-3.5}} & \multicolumn{2}{c}{\textbf{\textsc{Mistral-7B}}}&  \\
    \cmidrule(lr){3-4}\cmidrule(lr){5-6}
    && FL$\uparrow$ & CO$\uparrow$ & FL$\uparrow$ & CO$\uparrow$ & BooookScore$\uparrow$ \\
    \midrule
    \multirow{1}{*}{\textit{\rot{ABS}}}&HERCULES\textsubscript{ABS} &$4.00$&$1.64$&$4.20$&$2.16$& $39.56$ \\ \\
    \midrule
    \multirow{4}{*}{\textit{\rot{INC/HIE}}}&\textsc{Llama-3-8B-8k-Incremental} &$4.56$&$3.28$&$4.52$&$3.88$& $\underline{70.73}$\\
    &\textsc{Phi-3-Mini-3.8B-128k-Incremental} &$4.04$&$3.08$&$4.20$&$3.76$& $50.98$\\
    &\textsc{Llama-3-8B-8k-Hierarchical} &$\underline{4.64}$&$\underline{3.52}$&$4.44$&$\underline{4
    .08}$& $64.70$\\
    &\textsc{Phi-3-Mini-3.8B-128k-Hierarchical} &$4.32$&$3.36$&$4.44$&$4.00$& $55.59$\\
    \midrule
    \multirow{3}{*}{\textit{\rot{LLMs}}}&\textsc{Llama-3-8B-8k} &$4.60$&$3.12$&$4.48$&$3.72$&$67.71$ \\
    &\textsc{Phi-3-Mini-3.8B-4k} &$3.76$&$2.68$&$4.44$&$3.44$&$43.27$ \\ 
    &\textsc{Phi-3-Mini-3.8B-128k} &$4.47$&$3.19$&$4.36$&$3.44$& $57.06$\\ 
    \midrule
    \multirow{2}{*}{\textit{\rot{Ours}}}&\textsc{Xl-OpSumm(Phi-3-Mini-3.8B-4k)} &$\mathbf{4.68}$&$\mathbf{3.68}$&$\mathbf{4.72}$&$\mathbf{4.16}$& $66.23$\\ 
    &\textsc{Xl-OpSumm(Llama-3-8B-8k)} &$4.48$&$3.48$&$\underline{4.64}$&$\mathbf{4.16}$& $\mathbf{87.59}$\\
    \bottomrule
    \end{tabular}%
    }
    \caption{Reference-free evaluation on the \textsc{Xl-Flipkart} dataset. \textit{INC/HIE} indicates that the model uses either an Incremental or a Hierarchical approach. FL represents the average fluency score across all summaries generated by the model, while CO denotes the average coherency score. Refer to Appendix \ref{sec: metrics} for more description of these metrics.}
    \label{Table: reference_free_flipkart}
\end{table*}%

The evaluation of the generated summaries is conducted using the ROUGE-{1,2,L} F1 score (R1, R2 \& RL)\cite{lin-2004-rouge} and BERT-F1 score\cite{zhang2019bertscore}. Refer to Appendix \ref{sec: metrics} for more description of these metrics. It is noted that there is a possibility that the \textsc{Llama-3-8B-8k} and \textsc{Phi-3-Mini-3.8B-4k} models may not be able to handle the input tokens in \textsc{Xl-Flipkart} and AMASUM datasets. To address this, the input was truncated, and the maximum number of tokens that the models could handle was used to obtain the results. 

Table \ref{Table: automatic_eval_results} presents the results of various models on the AmaSum and \textsc{Xl-Flipkart} datasets. The analysis reveals the effectiveness of the \textsc{Xl-OpSumm} framework, particularly when employed with large language models (LLMs) such as \textsc{Llama-3-8B-8k} and \textsc{Phi3-3-Mini-3.8B-4k}.

On the AMASUM dataset, the \textsc{XL-OpSumm(Llama-3-8B-8k)} model outperforms its base and hierarchical variants across all metrics, including R1, R2, RL, and BERT-F1. It achieves the highest scores among all models for R1, RL, and BERT-F1, while being marginally outperformed by its incremental variant in terms of R2. Despite the \textsc{Phi3-3-Mini-3.8B-4k} model exhibiting higher ROUGE scores than the \textsc{Xl-OpSumm(Phi3-3-Mini-3.8B-4k)} model, the Wilcoxon Signed-Rank test indicates that the difference is not statistically significant.

On the \textsc{Xl-Flipkart} testset, the incremental variants of the \textsc{Llama-3} and \textsc{Phi-3} models outperform their corresponding base and hierarchical counterparts. Notably, when these LLMs are employed within the \textsc{Xl-OpSumm} framework, they surpass the performance of their incremental variants. Specifically, the \textsc{Xl-OpSumm(Llama-3-8B-8k)} model achieves the highest or second-highest scores across all metrics, outperforming the previous state-of-the-art models, such as HERCULES\textsubscript{EXT} and HERCULES\textsubscript{ABS}. 
\par The results demonstrate the effectiveness of the \textsc{Xl-OpSumm} framework in leveraging the capabilities of LLMs like \textsc{Llama-3-8B-8k} and \textsc{Phi3-3-Mini-3.8B-4k} for abstractive summarization tasks across diverse datasets like AMASUM and \textsc{Xl-Flipkart}. The framework consistently enhances the performance of these LLMs, enabling them to outperform existing state-of-the-art models.

\subsection{Reference Free Evaluation}
Traditional reference-based metrics like ROUGE inherently fail to capture the nuances of issues and contradictions within reviews, as demonstrated by prior work (\cite{bhaskar-etal-2023-prompted}, \cite{siledar2024prompt}). To address this limitation, we evaluate our framework across two dimensions: {\tt fluency (FL)} and {\tt coherence (CO)}(Appendix \ref{sec: metrics}), by prompting \textsc{GPT-3.5-Turbo} and \textsc{Mistral-7B-32k} models using the same method and prompts introduced in \citet{siledar2024prompt}. We could not evaluate the summaries on {\tt Relevance, Faithfulness, Aspect Coverage, Sentiment Consistency, Specificity} due to their input dependency and the token length limitations of the models under consideration (\textsc{GPT-3.5-Turbo} and \textsc{Mistral-7B-32k}). Additionally, we use {\tt BooookScore} \cite{chang2023booookscore} to evaluate the coherence of these summaries.

\noindent \textbf{\large AMASUM Dataset Evaluation} \\
Table \ref{Table: reference_free_amasum} presents the reference-free evaluation on the AMASUM dataset. All the LLM-based models outperform the HERCULES\textsubscript{ABS} model across all three metrics. Specifically, \textsc{Xl-OpSumm(Llama-3-8B-8k)} achieves the highest \textsf{avg}\footnote{\textsf{avg}: mean of scores given by \textsc{GPT-3.5} and \textsc{Mistral-7B} models as evaluators} {\tt Coherence} score of $4.04$ among its Llama-based variants, followed closely by \textsc{Llama-3-8B-8k-Hierarchical} with $4.02$. \textsc{Llama-3-8B-8k-Incremental} has an \textsf{avg} score of $3.94$. In terms of {\tt Fluency}, \textsc{Llama-3-8B-8k-Hierarchical} leads with an \textsf{avg} score of $4.86$, followed by \textsc{Xl-OpSumm(Llama-3-8B-8k)} with an \textsf{avg} score of $4.56$. In terms of {\tt BooookScore}, \textsc{Xl-OpSumm(Llama-3-8B-8k)} outperforms all other models with a score of $85.60$, followed by \textsc{Llama-3-8B-8k-Hierarchical} which achieved a score of $71.58$.

Among the PHI-3 models, \textsc{Xl-OpSumm(Phi-3-Mini-3.8B-4k)} excels with an \textsf{avg} {\tt Coherence} score of $3.98$ and an \textsf{avg} {\tt Fluency} score of $4.52$. It is closely followed by the \textsc{Phi-3-Mini-3.8B-128k-Hierarchical} model, which has \textsf{avg} scores of $4.5$ in {\tt Fluency} and $3.7$ in {\tt Coherence}. The \textsc{Phi-3-Mini-3.8B-128k-Hierarchical} model achieved the highest {\tt BooookScore} of $63.12$, closely followed by the \textsc{Phi-3-Mini-3.8B-128k} and \textsc{Xl-OpSumm(Phi-3-Mini-3.8B-4k)} models, which scored $61.86$ and $61.41$ respectively.

\noindent \textbf{\large \textsc{Xl-Flipkart} Dataset Evaluation} \\
Table \ref{Table: reference_free_flipkart} displays the reference-free evaluation on the \textsc{Xl-Flipkart} dataset. Models in the \textsc{Xl-OpSumm} framework outperform their Hierarchical and Incremental counterparts. \textsc{Xl-OpSumm(Llama-3-8B-8k)} achieves \textsf{avg} scores of $4.56$ in {\tt Coherence} and $3.82$ in {\tt Fluency}. The \textsc{Llama-3-8B-8k-Hierarchical} model scores an \textsf{avg} of $3.8$ in {\tt Coherence} and $4.54$ in {\tt Fluency}, while the \textsc{Llama-3-8B-8k-Incremental} model scores an \textsf{avg} of $3.58$ in {\tt Coherence} and $4.54$ in {\tt Fluency}. As observed in the AMASUM dataset, \textsc{Xl-OpSumm(Llama-3-8B-8k)} once again outperformed all other models, achieving a {\tt BooookScore} of $87.59$. This time, it was followed by \textsc{Llama-3-8B-8k-Incremental}, which scored $70.73$.

A similar trend is observed with the PHI-3-powered models. \textsc{Xl-OpSumm(Phi-3-Mini-3.8B-4k)} achieves the highest \textsf{avg} {\tt Coherence} score of $3.82$ and the highest \textsf{avg} {\tt Fluency} score of $4.7$ and a {\tt BooookScore} of $66.23$ among all PHI-3 models evaluated.

\subsection{Qualitative Analysis}
\label{sec: qualitative}
Table \ref{Table: qualitative} presents the summaries (Previous Global Summary, Local Summary, and Updated Global Summary) generated for a certain chunk of a Realme 8 product from the \textsc{Xl-Flipkart} dataset using \textsc{Xl-OpSumm (Phi-3-Mini-3.8B-4k)}. We observe that aspects such as build quality, Super AMOLED display, and gaming performance are new aspects present in the Local Summary. After referring to the aspect dictionary, aspects like the "display" and "battery life" of the mobile are updated in the global summary from the Previous Global Summary since they have the same sentiment in the aspect dictionary and Local Summary. For aspects like camera quality, there was dissatisfaction in the Previous Global Summary, but satisfaction concerning the back camera and dissatisfaction concerning the front camera in the Local Summary, so they are updated accordingly in the global summary referring to the aspect dictionary as well. Similarly, there was dissatisfaction in the Previous Global Summary for the aspect performance, but it was updated to a positive sentiment by referring to the Local Summary and aspect dictionary.

We also observed that specific information about aspects such as the MTK G95 processor model name, SONY IMX rear camera sensor, and 5000 mAh battery were dropped in some cases. Additionally, we observe a few hallucinations by the model, such as a rating of 4.5 out of 5 stars, which is not present in either the Local Summary or the Previous Global Summary.

\subsection{Comparative Analysis}
\label{sec: comparative}
Table \ref{Table: comparitive_analysis} shows summaries generated by various models for the Samsung Galaxy F23 5G. We observe that all the LLM-based summaries are coherent. However, the summary generated by HERCULES\textsubscript{ABS} lacks a structured overview and relevance to the product.

While other models successfully extract detailed information about the phone’s features, HERCULES\textsubscript{ABS} fails to do so. The Gold (GPT-4) summary stands out with its comprehensive coverage of multiple aspects, including display, battery life, performance, and camera quality, providing a balanced view highlighting both strengths and weaknesses.
The \textsc{Llama-3-8B-8k-Incremental} and \textsc{Xl-OpSumm(Llama-3-8B-8k)} summaries provide general overviews of user experiences but lack the depth and specific insights found in the Gold summary. The \textsc{Xl-OpSumm(Llama-3-8B-8k)} summary, in particular, highlights several positives not mentioned in the \textsc{Llama-3-8B-8k-Incremental}, such as the phone's durability and overall design quality. \textsc{Llama-3-8B-8k-Hierarchical} summary, while it is coherent, the length of the summary is very large compared to other model summaries.

\section{Summary, Conclusion and Future Work}
In this work, we introduce \textsc{Xl-OpSumm}, a scalable framework for opinion summarization that generates summaries incrementally from thousands of reviews. Additionally, we present a new test set, \textsc{Xl-Flipkart}, which contains thousands of reviews per product. Our framework can theoretically scale to process any number of reviews, regardless of the LLM context limit. Experimental results show that our framework outperforms all previous state-of-the-art models and other baselines on two datasets in ROUGE-based evaluations, and achieves higher average scores in reference-free evaluations across three dimensions.
\par
Studies from \citet{siledar2024product} showed that additional information sources are indeed helpful for opinion summarization task. Inspired from those works, a future direction for our work is to integrate additional sources such as Question-Answers and Product Descriptions into the \textsc{Xl-OpSumm} framework and to analyze their impact in the context of large volumes of reviews.

\section*{Limitations}
\begin{enumerate}
    \item Due to budgetary constraints associated with utilizing GPT-4, we have limited the Xl-Flipkart dataset to 25 products, for which we generated summaries using GPT-4. The principal objective of this study is to develop a framework capable of managing extensive contexts efficiently.
    \item We could not evaluate the summaries on {\tt Relevance, Faithfulness, Aspect Coverage, Sentiment Consistency,} and {\tt Specificity}. This is because these evaluations depend on the input and the models we used, \textsc{GPT-3.5-Turbo} and \textsc{Mistral-7B-32k}, have limitations on token length.
\end{enumerate}
\section*{Ethical Considerations}
While leveraging \textsc{GPT-4-Turbo} to generate summaries offers significant time and resource savings, we are aware of the potential impact on jobs related to summarizing and analyzing reviews. To address this, we are exploring methods to integrate human oversight with automated processes, striving to balance efficiency with job preservation.
Furthermore, users and stakeholders need to understand that these summaries are generated by AI. So we urge the research community to use the \textsc{Xl-Flipkart} test set with caution,
\bibliography{custom,anthology}
\appendix
\section{Global Summary Updation}
\label{sec: GSU}
When we process a chunk of reviews, they may have certain aspects that are not present in the previous chunks or may have information about the same aspects that conflict with the opinions from the previous chunks. Typically \textsc{Global Summary} represents important information from the previous chunks and the \textsc{Local summary} represents important information from the current chunk. When updating information in \textsc{Global Summary} using \textsc{Local Summary}, we handle the below 2 cases i.e. having new aspects in the \textsc{Local Summary} and conflicting opinions between \textsc{Local Summary} and \textsc{Global Summary} with the help of \textsc{Aspect Dictionary}


\noindent \textbf{a. New Aspects in the \textsc{Local Summary}}: In the case of a new aspect, we check the \textit{Aspect Dictionary} for the majority sentiment of that aspect. We only update \textsc{Global Summary} with new aspect information if the sentiment of that aspect in the \textsc{Local Summary} and the \textsc{Aspect Dictionary} matches. By doing this, we are making sure that the summary stays faithful to that aspect.

\noindent \textbf{b. Conflicting Opinions about an aspect between \textsc{Global} and \textsc{Local Summaries}}: In such cases, we again refer to the aspect's majority sentiment from the \textsc{Aspect Dictionary}. If it matches with sentiment in \textsc{Local Summary}, we update the \textsc{Global Summary} with the corresponding information from the \textsc{Local Summary}, else we leave the \textsc{Global Summary} as it is.

\noindent We embed all this information in a detailed prompt with a one-shot example and feed it to the LLM to update the \textsc{Global Summary}.


\begin{table}[hpt]
    \centering
    \resizebox{\columnwidth}{!}{%
    \begin{tabular}{p{\columnwidth}}
    \toprule
    \small{\textbf{Gold (GPT-4)}}\\
    \midrule
    \small{The Samsung Galaxy F23 5G offers a decent display, good battery life, and solid performance with the Snapdragon 750G at a competitive price. While the display is a TFT panel, it performs adequately for most users. The device's rear camera delivers satisfactory daylight images, but the front camera and low-light performance are average. Users appreciate the 120Hz refresh rate, although it experiences some inconsistency. Despite featuring a robust build and offering future-proof 5G compatibility, some customers have issues with fingerprint scanner reliability, adaptive brightness, and occasional software glitches. Notably, the phone comes without a charger, adding an extra expense for buyers. The device handles moderate gaming and day-to-day tasks well, with minor heat issues during extended use. While it includes Android 12 and promises good security features, users are concerned about the frequency and reliability of updates from Samsung. Overall, the phone is seen as a good, budget-friendly option, particularly for those prioritizing non-Chinese brands, despite some shortcomings in camera capabilities and accessory inclusion.}\\
    \midrule
    \small{\textbf{$\mathbf{HERCULES_{abs}}$}}\\
    \midrule
    \small{Will not power on. This camera is okay. Good performance. No more Cable bills! Idea. Tried multiple times. Did not need any Tech support. Doesn't hold a charge. Works fine. This thing is a beast. Very happy with this purchase. Good luck.}\\
    \midrule
    \small{\textbf{\textsc{Llama-3-8B-8k-Incremental}}}\\
    \midrule
    \small{Here is the updated summary:
The Samsung Galaxy F23 has received mixed reviews. While some praise its performance, camera, and display, others have reported issues with battery life, camera quality, and heating. The phone lacks a charger and charging cable in the box, a major drawback. Some users have reported issues with low-light camera performance, fast-draining battery, and auto brightness. However, the phone's display is good, and it has a smooth gaming experience. Overall, it's a good budget option, but may not meet premium device expectations.}\\
    \midrule
    \small{\textbf{\textsc{Llama-3-8B-8k-Hierarchical}}}\\
    \midrule
    \small{Here is the merged summary:

**Opinion Summary:**

The Samsung Galaxy F23 5G is a decent budget smartphone that offers a mix of pros and cons. While many users praise its performance, display, and camera quality, some have reported issues with battery life, with some experiencing rapid draining even with moderate usage. The phone's charger is not included in the box, which may be an inconvenience for some users.

The camera quality is a mixed bag, with some users praising it and others reporting poor performance in low-light conditions. The phone's 5G connectivity is a major selling point, although some users are disappointed that it does not support JIO 5G yet. Additionally, ...}\\
    \midrule
    \small{\textbf{\textsc{Xl-OpSumm (Llama-3-8B-8k)}}}\\
    \midrule
    \small{The Samsung phone has received mixed reviews. Users praise its camera quality, performance, and durability. However, some users have reported issues with battery drainage, overheating, and poor charging speed. Some users also experienced connectivity issues, sound quality issues, and poor network quality. The phone's display and design are also a subject of debate, with some finding it to be good, while others think it's average. Overall, opinions on the phone's value for money and performance vary, with some finding it a good budget option and others considering it a waste of money.}\\
    \bottomrule
    \end{tabular}%
    }
    \caption{\textbf{Comparative Analysis.} Summaries generated by various models on a product from the \textsc{Xl-Flipkart} dataset.}
    \label{Table: comparitive_analysis}
\end{table}

\section{Various Metrics Used in this Work}
\label{sec: metrics}
\textbf{ROUGE-1 (R1)}\cite{lin-2004-rouge} measures the overlap of unigrams (single words) between the generated summary and the reference summary. It gives an indication of how many individual words from the reference summary are captured in the generated summary. \\ \\
\textbf{ROUGE-2 (R2)}\cite{lin-2004-rouge} measures the overlap of bigrams (two consecutive words) between the generated summary and the reference summary. It provides insight into how well the generated summary preserves the sequence of word pairs from the reference summary. \\ \\ 
\textbf{ROUGE-L (RL)}\cite{lin-2004-rouge} calculates the longest common subsequence (LCS) between the generated summary and the reference summary. It captures the longest sequence of words that appear in both summaries in the same order, providing a measure of the overall structural similarity between the summaries.\\ \\
\textbf{BERT-F1}\cite{zhang2019bertscore} uses BERT, a pre-trained language model, to evaluate the similarity between the generated summary and the reference summary. BERTScore calculates precision, recall, and F1 score by comparing the contextual embeddings of words in both summaries, providing a more nuanced measure of semantic similarity than simple n-gram overlap.\\ \\ 
\textbf{FLUENCY (FL)}\cite{siledar2024prompt} assesses the quality of a summary in terms of grammar, spelling, punctuation, capitalization, word choice, and sentence structure. A fluent summary should be free of errors, and easy to read, follow, and comprehend. Annotators were given specific guidelines on how to penalize summaries based on their fluency levels.\\ \\ 
\textbf{COHERENCE (CO)}\cite{siledar2024prompt} evaluates the overall quality of the sentences in a summary. A coherent summary should be well-structured and well-organized, forming a logical and connected body of information rather than just a collection of related sentences.\\ \\
\textbf{\textsc{BooookScore}}\cite{chang2023booookscore} evaluates the coherence of summaries by prompting large language models (LLMs) to identify eight types of errors in each sentence. These errors include entity omission, event omission, causal omission, discontinuity, salience, language issues, inconsistency, and duplication. This metric is both reference-free and source-free.

\section{Other Baselines}
\label{sec: other_baselines} 
This section contains baselines that are not discussed in section \ref{sec: baselines}
\par 
\textbf{Oracle} represents the extractive upper bound computed by selecting input sentences with the highest R1 compared to the gold summary.
\par
\textbf{Random} represents selecting random reviews from the input as a lower bound.
\par \textbf{LexRank} \cite{Erkan2004LexRankGL} represents selecting the most salient sentences from the input by using BERT encodings to encode the sentences.
\par
\textbf{QT} \cite{angelidis-etal-2021-extractive} represents using vector quantization to map sentences to a discrete encoding space, then generates extractive summaries by selecting representative sentences from clusters. \par
 \textbf{SemAE} \cite{basu-roy-chowdhury-etal-2022-unsupervised} extends QT, relaxing the
discretization and encoding sentences as mixtures
of learned embeddings. 

\par \textbf{CopyCat} \cite{brazinskas-etal-2020-unsupervised} uses a hierarchical variational autoencoder that learns a latent code of the summary.

\section{Examples}
We present the example summaries of a few baselines, \xlopsummm{Llama-3-8B-8k} and \xlopsummm{Phi-3-Mini-3.8B-4k} in the Table \ref{Table: eg_summaries}.
\clearpage

\begin{table*}[t]
    \centering
    \resizebox{2\columnwidth}{!}{%
    \begin{tabular}{p{2\columnwidth}}
    \toprule
    \small{\textbf{Gold Summary}}\\
    \midrule
    \tiny{After evaluating numerous reviews, the Motorola G60 appears to be a robust mid-range smartphone offering consistent performance. Buyers have appreciated its large LCD display, smooth software experience with clean stock Android, and reliable battery life that comfortably lasts for over two days. The Snapdragon 732G chipset ensures lag-free gaming and smooth daily operation. While the device has a considerable weight and lacks an AMOLED screen, most users find these drawbacks negligible compared to the overall performance and build quality. The camera setup, though slightly slow in processing, provides satisfactory image quality in various settings. Critical points include the absence of stereo sound and fast charging, with the device taking around 2.5 hours for a full charge. Overall, the Moto G60 is considered a good value for its price, particularly praised for its user interface and build, making it suitable for users prioritizing functionality over high-end specifications.}\\
    \midrule
    \small{\textbf{HERCULES\textsubscript{ABS}}}\\
    \midrule
    \tiny{Very heavy. This camera does not disappoint. The battery lasts a long time. Doesn't feel cheap at all. Jim Does not come with a case. Worst product ever. Great sound and battery life. The battery last a long time. The sound quality is not that great. Very helpful. This camera is much better than the older model. Bought this for my wife for Christmas. Low light performance is great.}\\
    \midrule
    \small{\textbf{\textsc{Llama-3-8B-8k}}}\\
    \midrule
    \tiny{Here is a comprehensive opinion summary of the Motorola Moto G60:**Performance:** The phone has a good performance, with most users reporting smooth operation and no lag. **Display:** The display is a highlight of the phone, with many users praising its brightness, clarity, and refresh rate. **Camera:** The camera is a mixed bag, with some users reporting good quality images and others finding it to be below average. The front camera is generally praised, while the rear camera is criticized for its poor performance. **Battery:** The battery life is excellent, with many users reporting that it lasts for two days with moderate use. **Design:** The phone's design is a subject of debate, with some users finding it too bulky and others liking its sleek design. **Software:** The phone runs on stock Android, which is a major plus for many users. However, some users report occasional software issues and lag. **Value for money:** The phone is generally considered to be a good"}\\
    \midrule
    \small{\textbf{\textsc{Phi-3-Mini-3.8B-4k}}}\\
    \midrule
    \tiny{Based on these reviews, it seems like the sentiment towards the Motorola Moto G60 is mixed. Reviewers appreciate its display quality, camera performance, and value for money. However, they also express dissatisfaction with issues related to heating, camera functionality, and app performance, wh'}\\
    \midrule
    \small{\textbf{\textsc{Llama-3-8B-8k-Incremental}}}\\
    \midrule
    \tiny{Here is the updated summary: The Motorola G60 has received mixed reviews, with some users praising its performance, battery life, and display, while others have been disappointed with the camera quality. The phone's camera has been criticized for its poor low-light performance and some users have reported poor results with the selfie camera. The phone's display, battery life, and performance have been praised, but the camera quality is a major letdown. Despite this, the phone's performance is generally good, with no major lag or issues reported. The phone's battery life is also a highlight, with some users reporting up to 2.5 days of moderate use. However, the phone's charging time is slow and some users have reported issues with the camera app. Overall, the Motorola G60 is a good option for those looking for a budget-friendly device with good performance and battery life, but may not be the best choice for those prioritizing camera quality.}\\
    \midrule
    \small{\textbf{\textsc{Phi-3-Mini-3.8B-128k-Incremental}}}\\
    \midrule
    \tiny{Updated Summary:  Users have noted the Moto G60's sturdy construction, impressive battery life, and vibrant display, along with a solid stock Android experience. However, several users have criticized the phone for its slow camera, particularly in low light, heating issues during operation and charging, and s, and with with and and and and with. There. This such. And. and.s. and and and and and some for  and with withs. It and ands. and and the, and with and with and and like and and and and ands and and and and and and such such to such like like.  buts. This. like.s.s like such ands like such. like such. However. and and, and . with with with, like. and  and like. and and and  such like and such like like  such. with with such, and  like. and. However. However, but. Thiss. However withs.s, .s and to confirm, and. I. However.s.. And. There. and and. in ands ands. and. and. with with, and, and and with, a}\\
    \midrule
    \small{\textbf{\textsc{Llama-3-8B-8k-Hierarchical}}}\\
    \midrule
    \tiny{Here is the merged summary: The Motorola Moto G60 has received mixed reviews from customers, with both praise and criticism for its various features. On the positive side, many users appreciate its excellent battery life, good display, decent camera quality, and smooth performance. The phone's stock Android experience, fast charging, and good value for money are also notable strengths. However, some users have been disappointed with the camera quality, particularly in low-light conditions, and slow charging. Additionally, some have reported issues with hanging, overheating, poor camera performance, touch response, ghost touches, heating, and Wi-Fi connectivity problems. Some common praises include: Good performance and ...}\\
    \midrule
    \small{\textbf{\textsc{Phi-3-Mini-3.8B-128k-Hierarchical}}}\\
    \midrule
    \tiny{The Motorola G60 phone has garnered a mix of reviews, with users praising its commendable performance, long battery life, clean Android experience, compact design, and decent camera capabilities. Notably, the phone is appreciated for offering value for money within its price range. However, the reviews also consistently mention several areas of dissatisfaction. These include the display size and quality, particularly with the camera, in addition to software-related issues like heating during charging and app functionality. Speaker quality and the challenge in finding appropriate...}\\
    \midrule
    \small{\textbf{\xlopsummm{Phi-3-Mini-3.8B-4k}}}\\
    \midrule
    \tiny{The sentiment expressed in these reviews on the phone is mixed, with users appreciating the phone's performance, display, battery life, and clean UI. However, there are concerns about the heavy weight, poor camera quality (especially in natural light), and issues with the charging speed. Users also find the touch response to be slower than expected and note concerns about the phone's overheating during charging. Some users are satisfied with the phone's features and performance, while others find it not suitable for photography enthusiasts. The phone seems to offer good value for money, but it falls short in terms of camera and display quality compared to competitors. It is recommended for stock Android lovers and those prioritizing performance over photography. Overall, it's perceived as a great purchase in a big sale but advises caution due to some drawbacks."}\\
    \midrule
    \small{\textbf{\xlopsummm{Llama-3-8B-8k}}}\\
    \midrule
    \tiny{The Motorola Moto G60 has received mixed reviews, with some users praising its battery life, camera, and stock Android experience. However, many others have reported issues with the camera\'s performance, particularly in low-light conditions. Some users have also experienced heating problems, slow Wi-Fi connectivity, and poor customer service. Additionally, users have praised the phone\'s display quality and value for money. Overall, the phone seems to be a decent option for those looking for a budget-friendly device, but it\'s essential to set realistic expectations and weigh the pros and cons before making a purchase.}\\
    \bottomrule
    \end{tabular}%
    }
    \caption{Summaries generated by various models about Motorola G60 smart phone from \textsc{Xl-Flipkart} dataset. Gold Summary is summary generated by GPT-4-Turbo model.}
    \label{Table: eg_summaries}
\end{table*}%
\end{document}